\pdfoutput=1
\documentclass[runningheads]{llncs}

\usepackage{times}
\usepackage{mathtools}
\usepackage{array}

\usepackage[linesnumbered,ruled,vlined]{algorithm2e}
\include{pythonlisting}

\usepackage[T1]{fontenc}
\usepackage{amsmath}
\usepackage{url}

\usepackage{graphicx}
\usepackage{color}

\begin{document}
\title{Parallel Bayesian Optimization of \\Agent-based Transportation Simulation}
%
%

\author{Kiran Chhatre\inst{1}\thanks{Corresponding author: The work was performed as a Berkeley Lab affiliate.} \and
Sidney Feygin\inst{2} 
\and 
Colin Sheppard\inst{1,2} \and
Rashid Waraich\inst{1,2}}
\authorrunning{K. Chhatre et al.}

%
\institute{Lawrence Berkeley National Laboratory, Berkeley, CA 94720, USA\\
\email{\textbf{kiranchhatre3@gmail.com, \{colin.sheppard,rwaraich\}@lbl.gov}}\\ \and
Marain Inc., Palo Alto, CA 94306, USA\\
\email{\textbf{sid.feygin@gmail.com}}}
\maketitle              
\begin{abstract}
MATSim (Multi-Agent Transport Simulation Toolkit) is an open source large-scale agent-based transportation planning project applied to various areas like road transport, public transport, freight transport, regional evacuation, etc. BEAM (Behavior, Energy, Autonomy, and Mobility) framework extends MATSim to enable powerful and scalable analysis of urban transportation systems. The agents from the BEAM simulation exhibit ‘mode choice’ behavior based on multinomial logit model. In our study, we consider eight mode choices viz. bike, car, walk, ride hail, driving to transit, walking to transit, ride hail to transit, and ride hail pooling. The ‘alternative specific constants’ for each mode choice are critical hyperparameters in a configuration file related to a particular scenario under experimentation. We use the ‘Urbansim-10k’ BEAM scenario (with 10,000 population size) for all our experiments. Since these hyperparameters affect the simulation in complex ways, manual calibration methods are time consuming. We present a parallel Bayesian optimization method with early stopping rule to achieve fast convergence for the given multi-in-multi-out problem to its optimal configurations. Our model is based on an open source HpBandSter package. This approach combines hierarchy of several 1D Kernel Density Estimators (KDE) with a cheap evaluator (Hyperband, a single multidimensional KDE). Our model has also incorporated extrapolation based early stopping rule. With our model, we could achieve a 25\% L1 norm for a large-scale BEAM simulation in fully autonomous manner. To the best of our knowledge, our work is the first of its kind applied to large-scale multi-agent transportation simulations. This work can be useful for surrogate modeling of scenarios with very large populations.  

\end{abstract}
%
%
%

\section{Introduction}

In order to get from location ‘A’ to location ‘B’ we all choose a convenient mode of transportation. According to US Census Bureau \cite{census} study in 2017, people from New York metropolitan area spend 35.9 mins whereas people from the San Francisco bay area spend 32.1 mins in a typical one-way commute. Despite all the work transportation engineers put into making those journeys go smoothly, things go wrong more often than they should. While the weather plays an important role, one of the biggest reasons that transportation systems run into issues is that they have to deal with another difficult element, which is human behavior. In order to design a good transportation system, it is vital to understand the people who are going to use it.

Many people get around with vehicles like cars, ride-hail services, and bikes. According to Transport Statistics Great Britain study of 2017, 83\% \cite{trans_stat} of all journeys were taken by car, van or taxi. Potential collisions on roads aren’t confined to known junctions, moreover all back and forth lane switching causes congestion. Therefore, highway engineers tackle problems like this by developing traffic simulators. The forecasting of passenger travel requires multiple key elements that comprises the traffic simulator model. These key elements can be the estimation of trip generation (number of purposeful trips), trip distribution (destination choice), transportation mode choice, route assignment, trip chaining (the decision to link individual trips together in a tour), and several other aspects of traveler decisions \cite{wiki:trans}. 

The forecasting accounts for the prediction of specific transportation facility people will use in the future \cite{roadspacereq}. This forecast is dependent on the study of how people use transport. The factors affecting people’s behavior are broad and are based on activities and learning how people allocate their time during an average day. Forecasting begins with the collection of data on current traffic, population, employment, trip rates, travel costs \cite{predictnprovide}. Using this data a traffic demand model is developed, which returns estimate of the future traffic by feeding in predicted data like future population and employment. Such future traffic prediction is useful to calculate the capacity of infrastructure, to estimate financial and social feasibility of the projects, and calculate environmental impacts.

In our study, we propose algorithms for optimization of the “Mode Choice Analysis” in the traffic demand model. Mode choice analysis determines what mode of transport will be used and which modes of transport will be shared by the people. 

\begin{figure}[h]
\includegraphics[width=12cm]{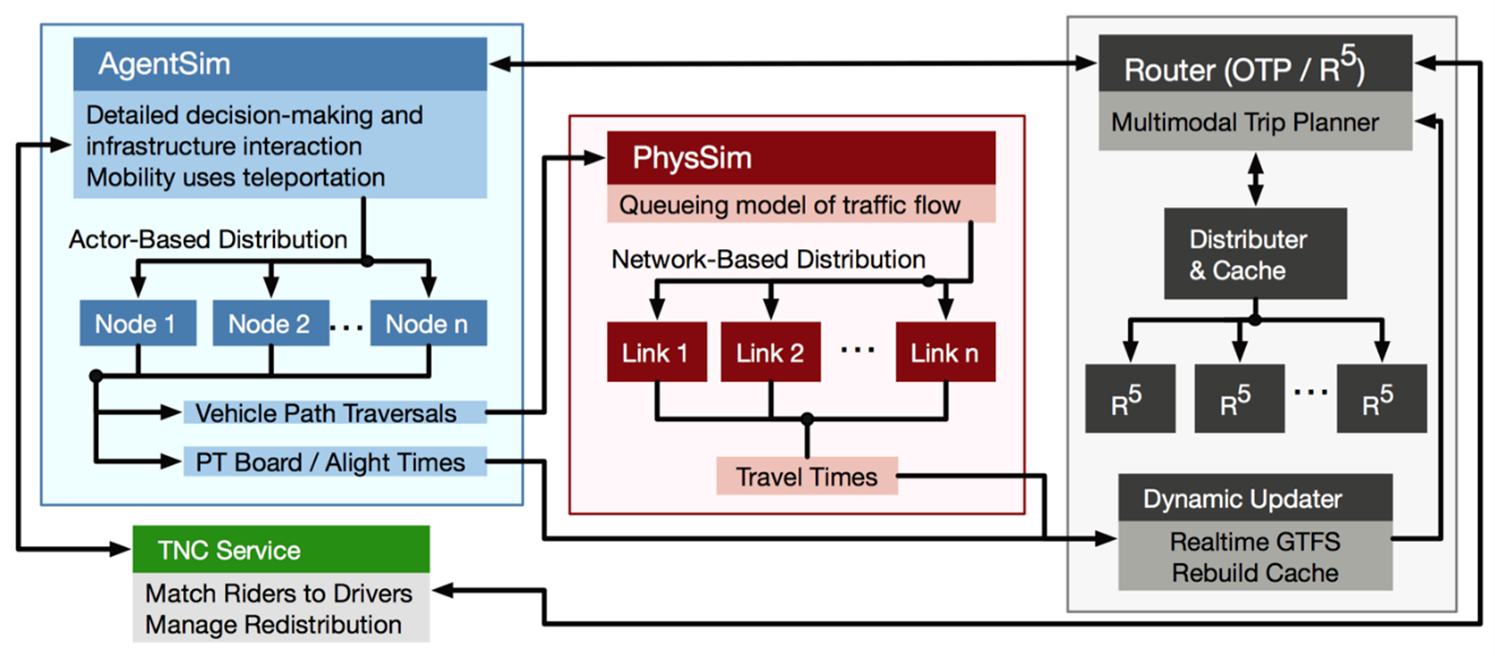}\label{f2}
\centering
\caption{BEAM Infrastructure \cite{beamdoc}}
\end{figure}

The large-scale transportation planning model of a particular whole city is executed through agent-based software modules. Complex systems have large number of interacting components that evolves over time. It’s difficult to predict the emergent behavior of such system at a macro level. On the other hand, even if we possess the knowledge of the macro structure, it’s difficult to find the microstructure that generates it. With the aid of computer-based modeling languages, we simulate complex patterns and understand more about how they arise in nature and society. In an agent-based model (ABM), the world is modeled using agents, the environment, and a description of agent-agent and agent-environment interactions, while an agent is an indivisible element with distinct properties and actions. Every agent is characterized as bounded rational and acts according to their perceived interests using decision-making rules in a predefined interaction topology. ABM abstracts the representation of a world that sometimes exaggerates certain aspects at the expense of others. However once the world is satisfactorily designed, the simulated experiments allows us to find out how agents interact with other agents that defines the patterns of behavior that are not defined at the level of any individual agent. With the help of ABM simulations we are able to explore how the individual’s micro decisions lead to macro patterns of the world’s behavior.

As shown in Figure \ref{f2}, BEAM \cite{osti_1398472} integrates MATSim \cite{Horni2016} toolkit and therefore possess an agent-based modeling approach. Each agent in BEAM employ reinforcement learning across successive simulated days to maximize their personal utility through plan mutation (exploration) and selecting between previously executed plans (exploitation). One of the primary objectives of the BEAM framework is to model and find an equilibrium point for all limited resource markets like road capacity, vehicle seating, transportation network companies (TNC) fleet availability, and refueling infrastructure that have limited supplies. Once the resources are utilized by travelers, no other traveler can simultaneously use the same resource. TNCs are modeled as a fleet of taxis controlled by a centralized manager that responds to requests from customers and dispatches vehicles accordingly. The degree to which an agent uses a resource depends on the resource availability and the agent’s behavior. In a particular case if supply of TNC drivers becomes limited, the wait time for hailing a TNC ride increases which in turn decreases the utility score of the TNC mode choice, and therefore reduced consumption of TNC resource. This dynamic choice process based on within day evaluation of modal alternatives allows the agent to maximize their utility. 

When BEAM is executed, the MATSim engine manages loading of demographic and network topology data as well as executes the BEAM simulator (BEAMMobSim), scores the metrics, and initiates the re-planning iterative loop. BEAM executes within-day planning that helps an agent govern their choice of transportation mode. A typical case would be an agent choosing whether to take a TNC ride after being reported with a wait time for a ride by the virtual TNC manager. Based on how long an activity was executed and how much time an agent spent traveling, BEAM re-planning loops are executed up to certain user-defined iterations (typically 15 iterations). This utility maximization through the re-planning process, unlike dynamic choice process based on within days evaluation, occurs outside the simulation day. Eventually, running BEAM over successive iterations balances the trade-offs between all resources in the system.

BEAMMobSim is composed of the AgentSim and the PhySim. AgentSim executes the daily plan of the population by allowing the agents to dynamically resolve the limited transportation resources. Whereas PhySim is a vehicle movement simulator. BEAM uses the R5 \cite{R5} routing engine to accomplish multi-modal routing. In the simulator, PhySim and AgentSim run serially. The agents receive the routing calculation (route and travel time) from the R5 router and choose between alternative routes, and later this information is used by PhySim to simulate traffic flow, resolve congestion, and update travel times back in the router. In the subsequent BEAM iterations, the agents move (teleport) according to travel times that are consistent with previous iteration network congestion.

The Metropolitan Transportation Commission \cite{MTC} (MTC) is the government agency responsible for regional transportation planning and financing in the San Francisco Bay Area. MTC coordinates transportation services in Bay area’s Alameda, Contra Costa, Marin, Napa, San Francisco, San Mateo, Santa Clara, Solano, and Sonoma counties. Vital Signs \cite{vital} is a data-driven web product developed by MTC that compiles data from a variety of national, state and regional data sources and makes it available to the public, agency staff, and policymakers to understand and track progress on key regional issues.

San Francisco Municipal Transportation Agency (SFMTA) undertook a Mode Share Survey within the San Francisco Bay Area using a survey conducted as a telephone study among 841 Bay Area residents aged 18 and older between May and August 2019 \cite{sfmta2019}. Mode share percentages based off the total number of trips (n = 10,437) for all respondents were determined by collecting trip level information for all respondents. In order to align with the scope of our study, “urbanism-10k” scenario was used. The urbanism-10k scenario is based on the City of San Francisco, including the SF Municipal public transit service and a sample population of 10,000 agents. 

The estimated mode share compiled by SFMTA and Vital Signs was adjusted to be suitable for the BEAM ubransim-10k mode choice model. The adjusted estimated mode share for the City of San Francisco (benchmark) is as Table \ref{table:modeshare}.

The intercept value for each mode choice affects the BEAM simulation in complex ways. Each intercept vector consists of eight distinct values that allows the AgentSim to reproduce realistic mode split travel behavior that is consistent with the data compiled by Vital Signs web service. Nonetheless, evaluating the accurate intercept vector is challenging due to inadequate prior knowledge of a bounded region that restricts the search of the mode choice model extrema. This concludes that it is essential to modify the intercept values in such a way that the simulation output converges. This process of manipulating the intercepts is termed as the calibration of the mode choice model. In that respect, BISTRO \cite{bistro} platform provides collaborative planning and evaluation of various optimization algorithms on agent-based modeling and simulation framework in response to different transportation policy strategies. 

\begin{table}
\caption{Mode Share for the City of San Francisco \cite{sfmta2019}}
\centering
\begin{tabular}{|cc|}
\hline
\textbf{Mode} & \textbf{Benchmark (\%)} \\
\hline
Bike & 2  \\
Car & 49 \\
Drive Transit & 4  \\
Ride Hail & 3 \\
Ride Hail Pooled & 2  \\
Ride Hail Transit & 1 \\
Walk & 22  \\
Walk Transit & 17 \\
\hline
\end{tabular}

\label{table:modeshare}
\end{table}

BEAM follows the MATSim convention for most of the input requirements to run a simulation and R5 convention for the road network and transit system inputs. Based on these external package requirements, BEAM requires population attributes, household population attributes, personal vehicle fleet, vehicle types for personal vehicles and the public transit fleet, R5 network and transit data, open street map network \cite{openstreet}, transportation analysis zone data, hereafter called "TAZ" \cite{wiki:trans}, and General Transit Feed Specification (GTFS) \cite{gtfs} data for each transit agency.


\section{State of the Art}

BEAM’s mode choice model is based on a multinomial logit choice model (MNL) in which agents select modal strategies (e.g. “car” versus “walk to transit” versus “TNC”) for each tour prior to the simulation day, but resolve the outcome of these strategies within the day (e.g. route selection, standard TNC versus pooled, etc.).

\subsection{Multinomial Logit Model}

In the conventional four-step transportation forecasting model, mode choice analysis comes the third after the trip generation step and the trip distribution step and is followed by the route assignment step. Trip distribution’s zonal analysis yields a set of origin destination tables describing where the trips will be made. Mode choice analysis allows the agents in the forecasting model to choose the mode of transport. The multinomial logit model (MNL) forms the backbone of the BEAM’s mode choice model. MNL proposes on making a choice of the heavier object from two objects with weights and greater the difference in weight, greater the probability of choosing correctly, where the perceived weight (\textit{u}) is 

\begin{equation}
u = v + e \;\;\;  \forall \;\;\;  e\in \mathcal{N}(\mu,\,\sigma^{2})
\end{equation}

And additionally, \textit{v} is the real weight and \textit{e} is a random variable independently and identically, normally distributed (having mean $\mu$  and variance $\sigma^{2}$ ) with \textit{v}. 

The economical root of MNL considers the additional characteristic of the object whose weight is being compared, converting the existing equation to 

\begin{equation}
u(x)=v(x)+e(x)  \quad  \text{and} \quad \log\left ( \frac{P_{i}}{1-P_{i}} \right ) = v(x_{i})
\end{equation}

where the second term is the MNL which is a log ratio of the probability of choosing a mode to the probability of not choosing the mode. The probability of choosing the mode depends on the real weight, in a general sense, the utility function of the mode choice. The higher the utility function, higher is the probability of choosing the mode choice. After a few algebraic manipulations to above equation, the probability value can be represented as

\begin{equation}
P_{i}=\frac{e^{v(x_{i})}}{1+e^{v(x_{i})}}
\end{equation}

and the utility function can be represented as

\begin{equation}
v(x_{i})=\beta_{i}+\beta_{cost}\cdot\text{cost}+\beta_{time}\cdot\text{time}+\beta_{transfers}\cdot\text{total transfers} \label{eq:4}
\end{equation}

where, \hfill \break
$\beta_{cost, time, transfers}$  = utility variable specific to the characteristic of the mode choice (i.e., cost, time and transfers) \\
$\beta_{i}$  = alternative specific constant (decision variable), it represents the characteristic that is not related to mode choice that includes additional nudge to make the random part of the equation independently normally distributed (NID)\\
cost, time, and total transfers = these are the mode choice specific parameters which are fixed values for every mode of transportation\\

The term $\beta_{i}$  is a mathematical value that primarily serves the purpose of transforming the random part of the above equation into an NID. However, the physical interpretation of the term can possibly denote an individual’s feeling of being politically correct by choosing a mode (e.g. Ride hail, with positive $\beta_{i}$ ) over other (e.g. Car, with negative $\beta_{i}$ ). This may additionally represent the individual’s justifiable decision with respect to their comfort, cost, and various other preferences. 

The utility \textit{v(x)} is not an observable parameter in the simulation. Rather the simulation output only specifies an observable response of whether a particular mode of transport was chosen (measured as 0 or 1). Rewriting the economical expression of choosing a mode  for all available eight choices by isolating the exogenous variables   (from the previous example, ‘real weight’ on an object) and their coefficients   (a vector of estimable parameters) at an observation \textit{t}, 

\begin{equation}
\vec{U}_{t}=\vec{V}_{t}\cdot\vec{\beta}+e_{t} 
\end{equation}

where $\vec{U}_{t}=[0,1]$ is a discrete response \\
and coefficients vector are $\:\beta_{i}^{j} (i\:\epsilon \left \{ cost, time,transfers\right \}$, $j\:\epsilon \left \{ 1,2,\cdots ,8\right \} )$

Therefore, the conditional probability $\Pr(U_{t}|V_{t})$  measures the chance that the observed outcome for choosing a particular mode of transportation is a noteworthy possible outcome for given exogenous variables, where  $\vec{\beta}$ describes the relationship between the observed and the real outcome. In this case the conditional probability takes the logistic form for two cases, first when the mode is selected and second when the mode is not selected:

\begin{equation}
    \Pr(U_{t}=1|V_{t})=\frac{e^{V_{t}\cdot\beta}}{1+e^{V_{t}\cdot\beta}} \quad \text{and} \quad \Pr(U_{t}=0|V_{t})=\frac{1}{1+e^{V_{t}\cdot\beta}}
\end{equation}

With this expression, we can calculate the estimator vector \cite{doi:10.1177/0049124102239083} $\vec{\beta}$  by maximizing the log-likelihood function $\ln\mathcal{L}(\beta)$. 

\begin{equation}
    \hat{\beta} =\operatorname*{arg\,max}_\beta [\ln\mathcal{L}(\beta)]
\end{equation}
\begin{equation}
    \hat{\beta} =\operatorname*{arg\,max}_\beta \left [ \sum_{t} \left ( U_{t} \ln\left ( \frac{e^{V_{t}\cdot\beta}}{1+e^{V_{t}\cdot\beta}}  \right )+ \left ( 1-U_{t} \right ) \ln\left ( \frac{1}{1+e^{V_{t}\cdot\beta}} \right )\right ) \right ]
\end{equation}

\subsection{Bayesian Optimization} \label{sec2.2}

Using Bayesian linear regression, we can infer the value of the function at point \textit{x} given the value of the function at point $x^{\prime}$. For a multivariate Gaussian prior distribution on vector $[f(x), f(x^{\prime})]$ with a mean function $\mu$  and a covariance function (‘kernel’) $\Sigma_{0}$, in the absence of enough data, we can compute the likelihood of the value of vector $[f(x), f(x^{\prime})]$ for a particular observation \cite{peterBayesian}. Due to Bayesian prior distribution setup, the section always occurs to be a univariate Gaussian distribution, the confidence interval for which using its standard deviation is computed using

\begin{equation}
    \bar{X}\pm Z\frac{\sigma }{\sqrt{n}}
\end{equation}

where  $\bar{X}$ is density mean, $\sigma$  is standard deviation, \textit{n} is sample size, and Z score for 95\% confidence is computed from the Z table \cite{zscore}. These computations when generalized over multiple dimensions, takes the form of Gaussian process regression as follows:

\begin{equation}
\begin{bmatrix}
f(x_{1})\\ 
\vdots \\
f(x_{k})
\end{bmatrix} \sim N \left ( \begin{bmatrix}
\mu(x_{1}) \\ 
\vdots \\
\mu(x_{k})
\end{bmatrix},\begin{bmatrix}
\Sigma_{0}(x_{1},x_{1})& \cdots  & \Sigma_{0}(x_{1},x_{k})\\
\vdots &\ddots  & \vdots\\ 
\Sigma_{0}(x_{k},x_{1}) & \cdots & \Sigma_{0}(x_{k},x_{k})
\end{bmatrix} \right)
\end{equation}

After creating a statistical inference model, with the help of a widely used acquisition function ‘Expected Improvement’ (EI) we can decide where to sample (Bayesian estimate) next. In this scenario, if we report the final solution after \textit{n+1} evaluations instead of \textit{n} evaluations, we would gain an improvement over global loss from \textit{F*} to min\textit{(F*,F(x))}. The benefit in the reduction of loss is given as  $EI(x)$ ,

\begin{equation}
    EI(x)=\mathop{E_{n}}\left [ F^{*} - min(F^{*},F(x)) \right ]=\mathop{E_{n}}\left [ min(F^{*},F(x))^{+} \right ]
\end{equation}

In the above form since we don’t know the \textit{F(x)}, the acquisition function uses the Bayesian posterior distribution of the objective to compute the conditional expectation of improvement of this probability distribution of \textit{F(x)} after ‘n’ evaluations. Therefore, the expected improvement  $EI_{n}(x)$ in closed form is evaluated using posterior mean and standard deviation in the formula:

\begin{equation}
    EI_{n}(x)=[\Delta_{n}(x)]^{+}+\sigma_{n}(x)\varphi \left ( \frac{\Delta_{n}(x)}{\sigma_{n}(x)} \right )-\left | \Delta_{n}(x) \right |\phi \left (- \frac{\left |\Delta_{n}(x)  \right |}{\sigma_{n}(x)} \right )
\end{equation}

where $ \Delta_{n}(x)=F^{*}_{n}-\mu_{n}(x)$, $\varphi$ = probability density function of posterior, and \\$ \phi$ = cumulative distribution function of posterior.

\subsubsection{Parallel Bayesian Optimization} 
If the problem’s domain allows multiple simultaneous evaluations, the setup explained above can be easily parallelized by parallelizing the EI in q points, where each point can consist of a multidimensional input vector. With parallel mode of evaluation, the Bayesian statistical model leverages diversity that aids in rapid convergence. The expression for EI is replaced by an unbiased gradient estimator $\nabla {EI}(x_{1:q})$ \cite{NIPS2016_6307}. By assuming Y to be a vector that consists of evaluations of the objective function from all q parallel workers, which also happens to be multivariate normal distribution under the Gaussian process posterior. In this setting, when we draw an independent standard normal vector Z, the expected improvement can be written as:

\begin{equation}
    Y=\left [ f(x_{1}),\cdots ,f(x_{q}) \right ]=m+CZ
\end{equation}

where mean $m = E\left [ Y \right ]$ and covariance  $C=\text{Cholesky(}\Sigma_{0}[Y]\text{)}$

\begin{equation}
   \dot{.\hspace{.095in}.}\hspace{.5in}\mathop{EI_{x_{1:q}}}=\mathop{E}\left [ \left ( F^{*}-min\left \{ Y \right \} \right )^{+} \right ]
\end{equation}

With this background, under satisfied regulatory considerations according to infinitesimal perturbation analysis theory \cite{HO199035}, we can switch derivative and the expectation as follows: 

\begin{equation}
    \nabla \mathop{EI_{x_{1:q}}}= \mathop{E}\left [ \nabla \left ( F^{*}-min\left \{ m+CZ \right \} \right ) ^{+} \right ] 
\end{equation}

Using the gradient estimator, we use a multi-start stochastic gradient method and iterate until global convergence. In the beginning, we start with random starting points equivalent to the number of parallel Bayesian workers, and using the gradient method, 

\begin{equation}
    \left ( \vec{x}_{1},\cdots ,\vec{x}_{q} \right )\leftarrow \left ( \vec{x}_{1},\cdots ,\vec{x}_{q} \right )+\alpha\nabla \mathop{EI_{x_{1:q}}} \quad \text{where} \;\alpha\; \text{is decaying step-size}
\end{equation}

With the decaying step size, the simulation progresses in an asynchronous and controlled fashion only in the direction where the largest gradient estimate is observed so that eventually the simulation converges to the stationary point of the $\nabla EI$  surface. The parallelized setup scales for large expensive-to-evaluate simulation for significantly large number of parallel workers each optimizing a input vector of up to twenty dimensions.

The prominent setback of Gaussian process with n evaluations is its cubic complexity $\mathcal{O}(n^{3})$  that comes from the inversion and determinant of the covariance kernel matrix $\Sigma_{0}$ . In contrast, Tree-structured Parzen Estimator approach (TPE) scales linearly in the number of evaluations. Unlike GP based Bayesian estimate $\Pr(Y|X)$  (integral of product of EI and posterior distribution), TPE estimate models  $\Pr(X|Y)$ and $\Pr(Y)$ separately. TPE defines $\Pr(X|Y)$  using two densities \cite{10.5555/2986459.2986743} 

\begin{equation}
\Pr(X|Y) =\left\{\begin{matrix}
l(x) \quad if \; y<y^{*}\\ 
g(x) \quad if \; y\geq y^{*}
\end{matrix}\right.
\end{equation}

where y denotes the objective loss value. Unlike GP based aggressive approach of choosing a single best served loss point, TPE chooses \textit{y*} that allows some points to be included in \textit{l(x)} such that 

\begin{equation}
    \Pr(y<y^{*}) =\gamma \quad \text{where} \; \gamma\; \text{is some quantile of observed \textit{y}}
\end{equation}

Therefore, the EI is computed as follows:

\begin{equation}
    \mathop{EI_{y^{*}}}(x)=\int_{-\infty}^{y*} (y^{*}-y)\Pr(y|x)dy \quad\text{and}\\
\end{equation}
\begin{equation}
\Pr(x)= \int_{R}\Pr(x|y)\cdot \Pr(y)\;dy=\gamma \cdot l(x)+(1-\gamma)\cdot g(x)\quad\text{so that}\\
\end{equation}
\begin{equation}
\mathop{EI_{y^{*}}}(x)\propto \left ( \gamma +\frac{g(x)}{l(x)}\cdot (1-\gamma ) \right )^{-1}
\end{equation}

The proportionality expression proves that the EI for evaluations is greater by minimizing the ratio $\frac{g(x)}{l(x)}$.

\subsection{Hyperband} \label{sec2.3}

While the objective function  $f:\chi\rightarrow \mathop{R} $ is expensive-to-evaluate, in BEAM simulation it is possible to define the cheap-to-evaluate approximate version $\tilde{f}(\cdot ,b)\; \text{of} \;f(\cdot )$  that are parameterized by budget-iters $b\;\in\; \left [ BEAM-lofi, BEAM-hifi \right ]$ . In our work, the budget \textit{b} is used to encode the number of iterations of the BEAM simulations. Hyperband \cite{li2016hyperband} is a parameter-free multi-armed bandit hyperparameter optimization algorithm that repeatedly calls the SuccessiveHalving (SH) \cite{pmlr-v51-jamieson16} method and is parameterized by different budgets \textit{b}. To identify the best configuration from \textit{n} randomly sampled configurations, Hyperband balances many aggressive runs on a smaller budget and very few runs that yields best results on maximum budget. For provided upper and lower limits of the budgets with a downsampling rate, total number of geometrically spaced SH stages are computed. For each stage, random configurations are drawn in a descending order where each configuration is allowed to run for computed iterations in an ascending order. For small to medium total budgets, Hyperband usually outperforms full function random search and Bayesian optimization evaluation methods. However, convergence to a global optimum is not guaranteed due to randomly drawn configurations.


\section{Proposed Method}

Our proposed method, as shown in Figure 2, parameterizes the optimization mechanism for BEAM such that it significantly reduces calibration times. Our method replaces current manual methods, that not only costs days for a subject matter expert to optimize, but also does not guarantee a desirable convergence. The optimization maximizes the accuracy of the model estimates and minimizes the cost associated with the parameterization. In this method, the algorithm approximates the real-world context of the BEAM simulation at lower iterations to estimate the margin of error on an extrapolated output at higher iterations. 

\begin{figure}[h]
\includegraphics[width=9cm]{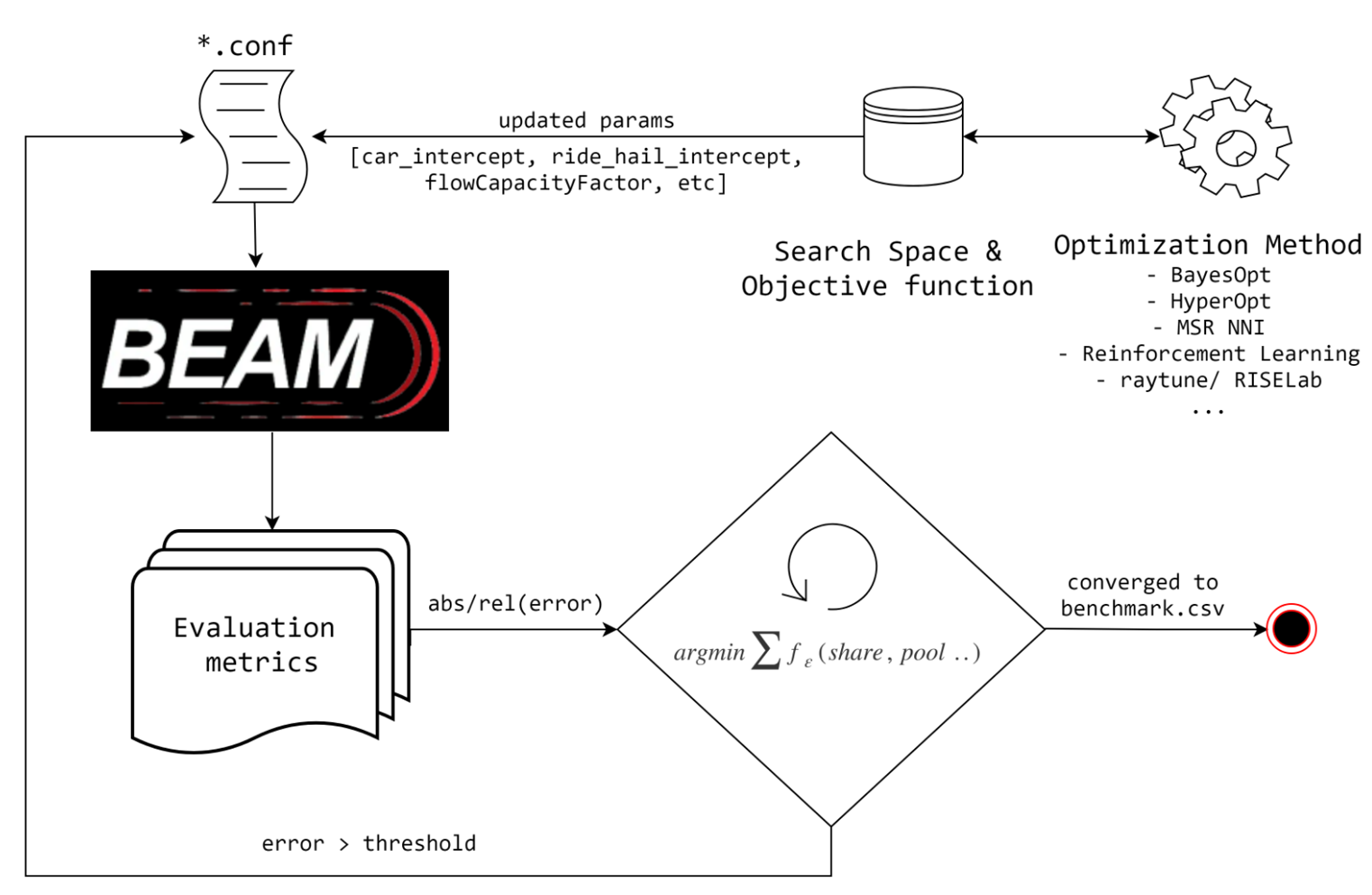}\label{fullmodel}
\centering
\caption{BEAM Optimization Model}
\end{figure}

\begin{algorithm}[H] \label{psuedo}
\SetAlgoLined

\textbf{Input: {\normalfont BEAM observations $Y$, ConfigSpace Input Variables $X$, budget-iters = [BEAM-lofi, BEAM-hifi]} } \\
\textbf{Output: {\normalfont $ x_{new}$}} \\
Initiate Pyro Name Server\\
Initial random $9$ BEAM runs: $D = \left \{ \left ( x_{0},y_{0} \right ),\cdots , \left ( x_{8},y_{8} \right ) \right \}$\\
$s_{max}=log_{\eta }\left [ \frac{BEAM-hifi}{BEAM-lofi} \right ]$\\
\For{$s\in \left \{ s_{max}, \cdots ,1,0 \right \}$ }{
Run HyperBand with $(\eta ^{s}\cdot BEAM-hifi)$ as initial budget: \\
Sample $x_{new} = \textup{max}\left ( \frac{l(x)}{g(x)} \right ) = \left ( \frac{p(y<\alpha |x,D)}{p(y>\alpha |x,D)} \right )$\\
Evaluate $y_{new} = BEAM(x_{new}) + \varepsilon $ \\
Refit $D \leftarrow D \cup (x_{new},y_{new})$
    }
 \caption{Bayesian Optimization with KDE and HyperBand}

\end{algorithm}

\subsubsection{Parallel Bayesian Optimization for Urbansim-10k Scenario} 
We used the Urbansim-10k scenario with a population of ten thousand agents for all optimization runs. The full BEAM run consists of fifteen iterations with an execution time of 3 hours on a machine with minimum system requirements with 8Gb RAM. However, for a vanilla version of Bayesian optimization setup, the acquisition functions suggest only one sample location in the search space. Therefore, to be able to run multiple workers updating the same posterior distribution through newer observations in an accelerated manner, we chose a parallelized Bayesian optimization setup. Parallelizing allowed us to choose which trials can continue to execute a full run of fifteen iterations and which trials can be pruned because of a bad performance. Additionally, the HyperBand integration in the HpBandSter \cite{hpbandster} package also provided us with another avenue to create a model-based early stop criterion. We use Gradle build automation tool for compilation of the BEAM code. Our Bayesian Optimization module focuses of high-fidelity optimization with an early stopping mechanism.

Based on the Section \ref{sec2.2} and Section \ref{sec2.3} the Python-based modules, BeamOptimizer and BeamWorker invoke the HpBandSter package to initiate the optimization task as described in Algorithm \ref{psuedo}. Additionally, configs and results are two JSON objects that book-keep the optimization progress of the model. The HpBandSter project is forked from its GitHub project (Commit: 841db4b) \cite{forked}. 

\subsubsection{BeamOptimizer}
BeamOptimizer manages the parallelization of the trials, where each trial is evaluating different configurations. It imports all the components from BeamWorker script and the Bayesian optimization algorithm from the HpBandSter project. The Pyro Name Server allows tracking all the threads in the network. We have created the experiments with the number of parallel threads ranging between four and sixteen. The script allows to arbitrarily choose the scenario that we want to optimize. We tested the optimization of BEAM's “Beamville” and the “Urbanism-10k” scenario. Followed by a scenario selection, our model generates unique copies of the scenario’s configuration file. Complete Bayesian optimization is run on a single Python process that initializes a Pyro \cite{pyro} Name Server.

\subsubsection{BeamWorker}
BeamWorker is based on the ConfigSpace \cite{configspace} package. ConfigSpace package manages and allows sampling of different types of hyperparameters. Each trial invoked by the BeamOptimizer is a copy of all procedures from the BeamWorker script. The optimizer algorithm executes the Successive Halving Algorithm (SH) (also known as HyperBand) during each iteration while simultaneously leveraging the new observation to update the posterior probability distribution. Each trial is associated with the iteration of the optimization algorithm, the budget of the current iteration, and the integer index of current trial.  The most compute expensive budget consists of 21 BEAM iterations. The objective function is the L1 norm between last iteration and the benchmark which is summed to generate a one-dimensional scalar value. This scalar value is returned to the Bayesian optimizer as the loss incurred after evaluating the drawn decision variables. All eight decision variables as a standard continuous float value are drawn from a specified range using the ConfigSpace package.

The early stop mechanism included in the BeamWorker script restricts the trail who’s L1 norm exceeds a predetermined value. The early-stopping module in our model determines the allowed highest L1 norm based on the total elapsed time since beginning the optimization experiment. The L1 threshold value decreases from 115\% to 5\% from the first elapsed 150 minutes until 750 minutes. This threshold value is dynamically imported in the BeamWorker script and is compared against the intermittent L1 norm at the end of third BEAM iteration. 

\begin{table}
\caption{Parallel Bayesian Optimization on BEAM Urbansim-10k scenario. The results demonstrate L1 norm improvement for the BEAM's mode choice model on the specified configuration space. The configuration space is a hyperparameter for the optimization task.}\label{tab1}
\begin{tabular}{|c|c|c|c|c|c|c|}
\hline
Optimization-runs &  BEAM iters & Parallel Worker & High L1 \% & Low L1 \% & ET hrs & ConfigSpace \\
\hline
10 & 21 & 8 & 183 & 60 & 6.5 & [-20,20] \\
25 & 21 & 16 & 171 & 70 & 7 & [-20,20] \\
40 & 21 & 4 & $\approx$180 & 61 & -- & [-20,20]\\
25 & 21 & 4 & 200 & 48 & 28 & [-100,100] \\
50 & 21 & 5 & 40 & \textbf{25} & 51.5 & Optimal $\pm$ 5\%\\
100 & 21 & 16 & 160 & \textbf{25 (40\textsuperscript{th} iter)}  & $\approx$ 72 & Optimal $\pm$ 20\% \\

\hline
\end{tabular}
\end{table}

\subsubsection{Computing Platform}

All experiments were conducted on Amazon Elastic Compute Cloud (EC2). For our project, we used a general purpose EC2 instance as these instances uses the resources in equal proportions for any code repository. We used a m5a.24xlarge instance for all experiments. This instance provided 96 vCPU (with 384 GB of computer memory), where each vCPU is a single physical CPU core on EC2 instance’s operating system.

\section{Experiments}

\begin{figure}[h]
\includegraphics[width=10cm]{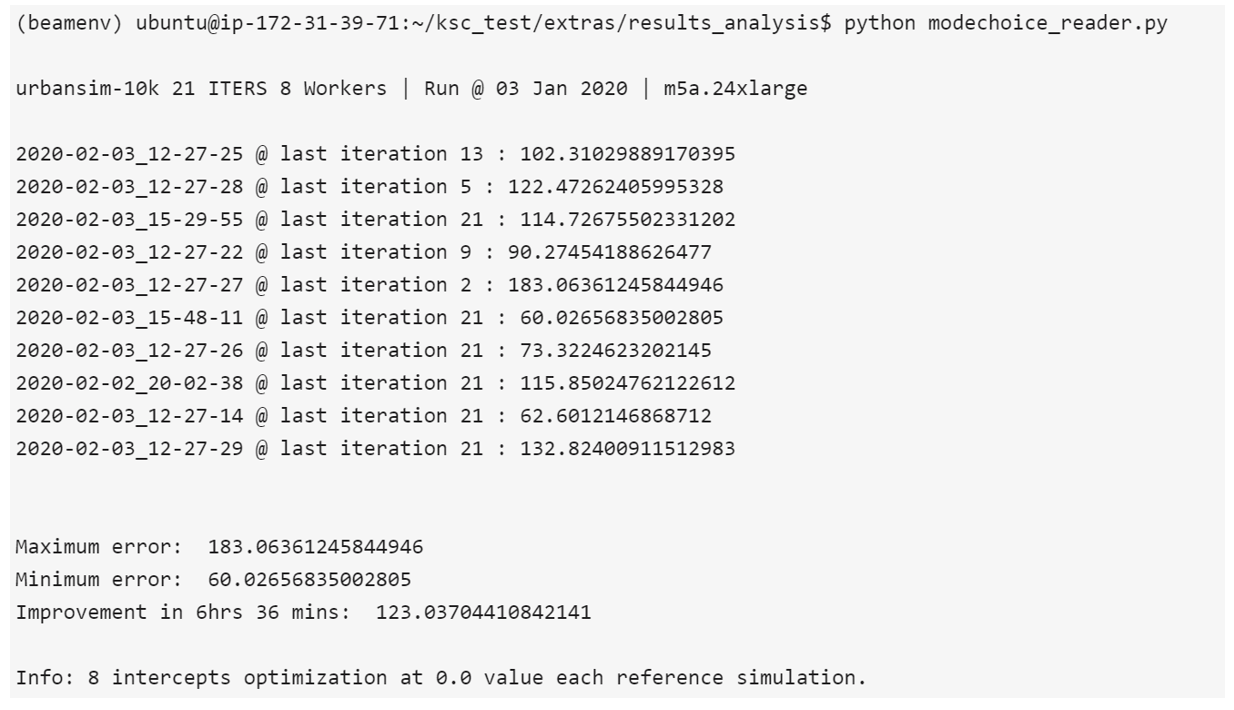}
\centering
\caption{8 Parallel Workers and 21 BEAM Iterations} \label{expss}
\end{figure}

\begin{figure}[h]
\includegraphics[width=7.5cm]{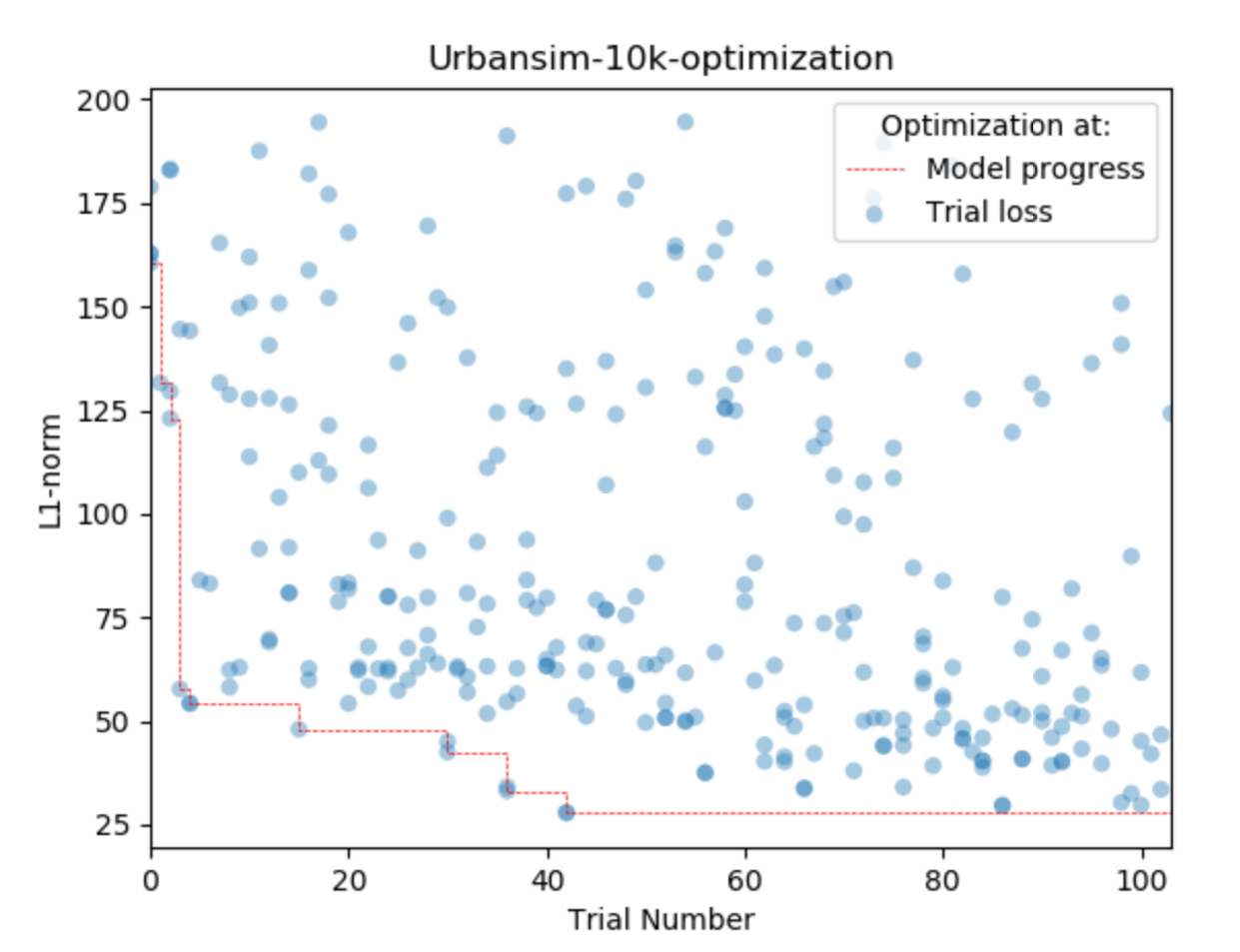}\label{best}
\centering
\caption{16 Parallel Workers and 100 Optimization Iterations}
\end{figure}

We conducted optimization experiments to validate the performance of our optimization routine on the BEAM Urbansim-10k scenario, a selective list is denoted in Table \ref{tab1}. The best score that we obtained through the Bayesian optimization model is 25\% L1 norm. We varied the number of parallel workers and BEAM iterations to observe the optimization performance on the BEAM's mode-choice objective function. Figure \ref{expss} represents a sample output of one of our early experiments with objective loss of 60\%. The total number of BEAM iterations for this run were 21 and the model executed with eight parallel workers. The maximum L1 norm was observed to be 183\% and the experiment ran up to 6.5 hours wall clock time. We observed the early stop operation at fewer trials being executed later (fifth, nineth) than the specified third BEAM iteration.

Followed by this experiment, we increased the number of parallel workers and number of optimization iterations to 16 and 25 respectively. The model showed an improvement from 172\% L1 norm to 70\% L1 norm over up to 7 hours of optimization wall clock time. In order to improve the frequency of at which the Bayesian posterior distribution was being updated, we then conducted an experiment with high optimization iterations (40) and less parallel workers (4). In this case, the lowest reported score was 61\% L1 norm.

All experiments previously mentioned had a small search space range [-20,20] from where the ConfigSpace package would search and sample the optimal decision variable value. To verify the effect of widening the search space, we now expanded the search space between [-100,100] and conducted longer experiments out of which one lasted for 28 hours wall clock time with 25 optimization iterations. The model started from highest score close to 200\% L1 norm and converged up to 48\% L1 norm. We conducted additional experiment with narrower search spaces that represented a region of 5\% variance around the optimal intercept values discovered from the manual calibration experiment. This experiment included 16 parallel workers that led to 100\% AWS EC2 CPU utilization with a maximum of 16 GB of memory usage per BEAM trial. We conducted total of 50 optimization iterations that lasted for 51.5 hours of wall clock time. Due to a narrow search space, the model started off well at 40\% L1 norm but could improve up to 25\% L1 norm by the end of the experiment.

Moreover, we conducted very large scaled optimization experiments consisting of 100 optimization iterations with a maximum of 21 BEAM iterations per trial and 16 parallel workers as shown in Figure \ref{best}. The experiment constantly utilized 100\% EC2 CPU memory and lasted for more than three consecutive days. The search space was not manipulated around the ideal intercept values and had a variance of 20\% around 0-mean value. This experiment was recorded to deliver the best performance of the Bayesian optimization model that costed about 2500 BEAM Urbansim-10k iterations. The experiment yielded an improvement from 160\% L1 norm to 25\% L1 norm at close to 40th optimization iteration. The optimization improvement tend to level out after the 45th optimization iteration.

\section{Conclusion and Future Directions}

We present an optimization model at scale to calibrate a multi-agent transportation Urbansim-10k BEAM simulation. Given a particular distributional forecast of demand patterns tuned to the particular time period, the BEAM simulation estimates the real time assignment decisions for agents to their mode-choice requests. In our study, we have focused on optimizing the intercepts for car, walk, walk transit, drive transit, ride hail transit, ride hail, ride hail pooled, walk transit, and bike parameters. All these parameters in totality contribute towards the mode choice model’s convergence. In order to accelerate the convergence of such simulator, we developed a parallel Bayesian optimization model. Using our model, with a 20\% variance search space around the 0-mean optimal value, we executed about 2500 BEAM iterations to achieve an L1 norm improvement from 160\% to 25\% using a fully automated procedure. The measured performance was independent of the chosen scenario and the BEAM’s “\textit{lastIteration}” value. 

The ultimate usage of this optimization tool is planned to be made on the full San Francisco Bay Area population with all objectives applied. Alternative research questions can be formulated on this front. Research questions with broader context like multi-task learning and warm-start can be sought while considering all the input parameters. Extending the learned prior distribution of an experiment to geographically dissimilar scenario is also a problem of increasing importance.

\subsubsection{Acknowledgements} The authors would like to thank Artavazd Balaian, Haitam Laarabi, Nikolay Ilin, Jessica Lazarus, Zachary Needell, and Rajnikant Sharma for many insightful conversations and technical support. The research was funded by Berkeley Lab's fellowship and the German National Scholarship provided by the Hans Hermann Voss Foundation.

%
%
%
%

\end{document}